\pdfoutput=1

\documentclass[11pt]{article}

\usepackage{EMNLP}

\usepackage{xurl}
\usepackage{xspace}
\usepackage{times}
\usepackage{latexsym}
\usepackage{smartdiagram}
\usepackage{tikz}
\usepackage{tcolorbox}
\usetikzlibrary{fit, arrows.meta,
                positioning
                }
\usetikzlibrary{shapes, shadows}
\usepackage{threeparttable}
\usepackage{graphicx}
\usepackage[T1]{fontenc}

\usepackage[utf8]{inputenc}
\usepackage{enumitem}
\usepackage{microtype}
\usepackage{amsmath, mathtools}
\usepackage{pgfplots} 
\pgfplotsset{width=7cm,compat=1.8} 
\usepgfplotslibrary{polar}
\usepackage{pgfplotstable} 
\usetikzlibrary{
        calc,
        matrix,
    }
\usepackage{multirow}

\newcommand{\qaid}[1]{
\centering{
\begin{tcolorbox}[colback=orange!20, colframe=white, left=0pt, right=0pt, top=0pt, bottom=0pt, width=0.3\textwidth]
\centering{\tn{#1}}
\end{tcolorbox}}
}
\newcommand{\qabox}[2]{
\centering{
\begin{tcolorbox}[colback=#1!10, colframe=white, left=0pt, right=0pt, top=0pt, bottom=0pt]
#2
\end{tcolorbox}}
}
\newcommand{\gptbox}[2]{
\centering{
\begin{tcolorbox}[colback=#1!7, colframe=white, left=0pt, right=0pt, top=0pt, bottom=0pt]
{\color{#1} #2}
\end{tcolorbox}}
}
\usepackage{inconsolata}
\usepackage{xcolor}
\usepackage{amsthm, amsmath, amsfonts}
\usepackage{bm}

\tikzset{
  heading/.style={
    rectangle,
    rounded corners,
    draw=gray,
    align=center,
    minimum width=8em,
    text width=8em,
    font=\scriptsize,
    fill=orange!70!yellow!30,
    top color=orange!70!yellow!50,
    bottom color=orange!70!yellow!10,
    drop shadow={shadow xshift=2pt, shadow yshift=-2pt, fill=gray!70, opacity=0.6}
  },
  example_heading/.style={
    rectangle,
    rounded corners,
    draw=gray,
    align=center,
    minimum width=8em,
    text width=10em,
    font=\scriptsize,
  },
  example_prompt/.style={
    rectangle,
    rounded corners,
    draw=gray,
    align=center,
    minimum width=8em,
    text width=38.5em,
    font=\scriptsize,
  },
  inline_prompt/.style={
    rectangle,
    rounded corners,
    draw=gray,
    align=left,
    minimum width=18em,
    text width=18em,
    font=\scriptsize,
  },
prompt/.style={
    rectangle,
    rounded corners,
    draw=gray,
    align=left,
    minimum width=18em,
    text width=18em,
    font=\scriptsize,
    fill=cyan!70!blue!20,
    top color=cyan!70!blue!30,
    bottom color=cyan!70!blue!10,
    drop shadow={shadow xshift=2pt, shadow yshift=-2pt, fill=gray!70, opacity=0.6}
  },
  response/.style={
    rectangle,
    rounded corners,
    draw=gray,
    align=center,
    minimum width=18em,
    text width=18em,
    font=\scriptsize,
    fill=magenta!50!red!20,
    top color=magenta!50!red!30,
    bottom color=magenta!50!red!10,
    drop shadow={shadow xshift=2pt, shadow yshift=-2pt, fill=gray!70, opacity=0.6}
  },
  header/.style={
    rectangle,
    align=center,
    minimum width=18.5em,
    text width=18.5em,
    font=\scriptsize,
    fill=magenta!50!red!20,
    top color=magenta!50!red!30,
    bottom color=magenta!50!red!10,
  },
ultraminiprompt/.style={
    rectangle,
    rounded corners,
    draw=gray,
    align=left,
    minimum width=11.3em,
    text width=11.3em,
    font=\scriptsize,
    fill=#1!70!blue!20,
    top color=#1!70!blue!30,
    bottom color=#1!70!blue!10,
    drop shadow={shadow xshift=2pt, shadow yshift=-2pt, fill=gray!70, opacity=0.6}
  },
  miniprompt/.style={
    rectangle,
    rounded corners,
    draw=gray,
    align=left,
    minimum width=17em,
    text width=17em,
    font=\scriptsize,
    fill=#1!70!blue!20,
    top color=#1!70!blue!30,
    bottom color=#1!70!blue!10,
    drop shadow={shadow xshift=2pt, shadow yshift=-2pt, fill=gray!70, opacity=0.6}
  },
  miniresponse/.style={
    rectangle,
    rounded corners,
    draw=gray,
    align=center,
    minimum width=17em,
    text width=17em,
    font=\scriptsize,
    fill=#1!50!red!20,
    top color=#1!50!red!30,
    bottom color=#1!50!red!10,
    drop shadow={shadow xshift=2pt, shadow yshift=-2pt, fill=gray!70, opacity=0.6}
  },
miniheader/.style n args={2}{
    rectangle,
    align=center,
    minimum width=16.65em,
    text width=16.65em,
    font=\scriptsize,
    fill=#1!50!#2!20,
    top color=#1!50!#2!30,
    bottom color=#1!50!#2!10,
  },
  arrow/.style={
    ->,
    draw=gray,
    line width=2mm,
    shorten >=1pt,
    shorten <=1pt,
    -{Triangle[scale=0.6]}
  }
}

\newcommand{\addHeader}[5]{
        \node [miniheader={#3}{#4}, right,inner xsep=1em,outer sep=0pt,text height=2ex,text depth=.5ex] (#2)
            at ([shift={(-0.5em,-0.3em)}]#1.north west) {#5};
        \fill[miniheader={#3}{#4}] (#2.north east) -- +(-0.5em,0.25em) -- +(-0.5em,0) -- cycle;
        \fill[miniheader={#3}{#4}] (#2.south west) -- +(0.5em,-0.25em) -- +(0.5em,0) -- cycle;
}

\newcommand{\scr}[1]{{\scshape #1}}

\definecolor{deepblue}{HTML}{1f77b4}
\definecolor{lightred}{HTML}{ffcccb}
\definecolor{vibrantorange}{HTML}{ff7f0e}
\definecolor{lushgreen}{HTML}{2ca02c}
\definecolor{elegantpurple}{HTML}{9467bd}
\definecolor{olivegreen}{RGB}{128, 128, 0}
\definecolor{darkgreen}{RGB}{0, 100, 0}
\newcommand{\Phimn}{\scr{phi-3-mini}\xspace}
\newcommand{\Phimd}{\scr{phi-3-medium}\xspace}
\newcommand{\Orca}{\scr{Orca-2}\xspace}
\newcommand{\Mistral}{\scr{Mistral}\xspace}
\newcommand{\fPhi}{\scr{phi-3}\xspace}
\newcommand{\sPhi}{\scr{Phi}\xspace}
\newcommand{\sOrca}{\scr{Orca}\xspace}
\newcommand{\sMistral}{\scr{Mstl}\xspace}
\newcommand{\gpt}{\scr{GPT-4}\xspace}
\newcommand{\cgpt}{\scr{GPT-3.5 Turbo}\xspace}
\newcommand{\scgpt}{\scr{GPT-3.5T}\xspace}

\newcommand{\clb}[1]{{\color{blue} #1}}

\newcommand{\cldg}[1]{{\color{darkgreen} #1}}
\newcommand{\clm}[1]{{\color{magenta} #1}}
\newcommand{\clo}[1]{{\color{orange} #1}}

\newcommand\tn[1]{\textbf{\small #1}}

\title{Fine-tuning Smaller Language Models for Question Answering over Financial Documents}

\author{
\bf{Karmvir Singh Phogat,} 
\bf{Sai Akhil Puranam,}
\bf{Sridhar Dasaratha,}\\ 
\bf{Chetan Harsha,}  
\bf{Shashishekar Ramakrishna}
\\
EY Global Delivery Services India LLP
\\
\texttt{\{Karmvir.Phogat,Sai.Puranam,Sridhar.Dasaratha\}@gds.ey.com},\\
\texttt{\{Chetan.Harsha,Shashishekar.R\}@gds.ey.com}
}

\begin{document}
\maketitle

\begin{abstract}
Recent research has shown that smaller language models can acquire substantial reasoning abilities when fine-tuned with reasoning exemplars crafted by a significantly larger teacher model. We explore this paradigm for the financial domain, focusing on the challenge of answering questions that require multi-hop numerical reasoning over financial texts. We assess the performance of several smaller models that have been fine-tuned to generate programs that encode the required financial reasoning and calculations. Our findings demonstrate that these fine-tuned smaller models approach the performance of the teacher model.

To provide a granular analysis of model performance, we propose an approach to investigate the specific student model capabilities that are enhanced by fine-tuning. Our empirical analysis indicates that fine-tuning refines the student models ability to express and apply the required financial concepts along with adapting the entity extraction for the specific data format. In addition, we hypothesize and demonstrate that comparable financial reasoning capability can be induced using relatively smaller datasets.
\end{abstract}

\section{Introduction}
In recent years, the development of large language models (LLMs) has achieved significant advances in natural language processing (NLP). Scaling up the size of these models has not only improved sampling efficiency and performance, \cite{scaling_laws} but also introduced reasoning capabilities \cite{cot,zs_cot}. LLMs have been shown to perform well on tasks requiring reasoning capabilities in various domains, including code writing \cite{evaluating_llm_code}, math problem solving \cite{solving_qr, formal_mathematics}, dialogue \cite{improving_alignment,lamda}, common sense reasoning \cite{unsupervised_commonsense, palm} and symbolic reasoning \cite{emergent_abilities,sc_cot}.

A major drawback of these methods is their reliance on extremely large models with hundreds of billions of parameters. These models can be costly to deploy at scale due to high computational requirements and inference costs. To overcome these limitations, recent research has focused on inducing reasoning in smaller models. A common approach is to use large models to generate training samples with demonstrations of reasoning on one or more tasks that are then used to fine tune smaller models \cite{teaching_slm,reasoning_teachers}. While these methods have shown promising results on various tasks including arithmetic, symbolic and common-sense reasoning, the applicability, and effectiveness of these methods in specific domains such as finance need further exploration. 
Question answering in the finance domain poses a unique set of challenges, requiring the understanding of financial concepts along with the ability to perform numerical reasoning. This complexity introduces a significant challenge that is distinct from classical question answering problems \cite{hotpotqa,know_what}

In this paper, we present an empirical study that provides experimental evidence supporting the effectiveness of fine-tuning small language models for financial question answering. Our research is guided by several critical questions:

RQ1: To what degree does fine-tuning small language models improve their performance on financial question answering tasks?

RQ2: What are the intrinsic characteristics of the base language model that contribute to its performance prior to fine-tuning?

RQ3: Which aspects of question answering benefit directly from the pre-trained knowledge, and what specific improvements are enabled by fine-tuning?

RQ4: What are the fine-tuning data requirements to achieve these improvements?

To address these questions, we adapt previous approaches on inducing reasoning in smaller models for the financial question answering task. In our experimental setup, we employ \gpt as the teacher model, building upon its documented success in the realm of financial question answering \cite{pot,zspot}. For the student models, we explore a suite of state-of-the-art, yet relatively smaller, language models including \fPhi variants (3.5B and 14B parameters) \cite{phi3}, \Mistral 7B, and \Orca configurations (7B and 13B parameters) \cite{orca2}. Our methodology involves training the student model using Python programs generated by the teacher model. The teacher generated code systematically delineates the steps for financial reasoning, including question comprehension, formula identification and entity extraction. At inference time, the language models are tasked with producing Python code, which is then executed by an external interpreter. 
This code generation and external execution strategy has been demonstrated to be more effective than methods that rely on the language model to internally perform the calculations.
\cite{pal,pot,zspot}.

Our main contributions are summarized as follows:
\begin{enumerate}[leftmargin=*, widest=b, align=left]
\item We refine the process of fine-tuning smaller language models by introducing an approach designed for question answering over financial reports. 
\item Our experimental study on three financial datasets provides insights on the performance of the small models
\item We propose an evaluation method that utilizes \gpt to assess the Python code outputs from the student models, pre- and post-fine-tuning. Combining \gpt's automated assessment with manual evaluation yields new insights into the distinct competencies that are enhanced in the student model during fine-tuning.
\item Motivated by these insights we explore the use of smaller datasets for fine-tuning and provide empirical evidence demonstrating their effectiveness.
\end{enumerate}

Our experimental findings reveal that smaller language models fine-tuned for financial reasoning, can achieve performance that rivals that of the larger teacher model. Moreover, our empirical analysis suggests that the fine-tuning helps refine concept understanding and enables consistent reasoning with those concepts. In addition, the entity extraction abilities get honed on the specific format of the financial dataset.

\section{Background}
Pre-trained large language models are shown to perform well on tasks requiring reasoning when used with certain techniques. \cite{cot} proposed prompting the LLMs to solve the problem step-by-step by providing a few exemplars. \cite{pot} proposed a few-shot prompt to produce a program which is then executed externally. However, methods relying on pre-trained LLMs can be costly to deploy at scale.

Recent efforts attempt to replicate these reasoning capabilities in small language models. One of the common approaches is following a teacher-student setup where a pre-trained LLM acts as a teacher, generating training data which is used to teach a small language model, the student. \cite{orca,orca2} aim to train models to exhibit generic reasoning abilities. They utilize \cgpt and \gpt as teacher models to generate training data with carefully crafted prompts. On the other hand, \cite{msr,teaching_slm,reasoning_teachers} train task specific small language models with CoT based explanation from pre-trained LLMs. Specifically for problems involving mathematical reasoning, \cite{mathcoder,tora,openmath_instruct,codet5} propose to generate programs from the pre-trained LLMs and train the small language models. In contrast, we focus on fine-tuning small language models for financial question answering problems \cite{finqa,convfinqa,tatqa}. Solving these problems requires financial domain knowledge and complex reasoning compared to the math word problems addressed in previous studies.

Prior works have studied the use of pre-trained LLMs for financial question answering. \cite{assess_fqa} perform a detailed comparison of the performance of pre-trained LLMs on financial datasets with various prompting techniques. They also introduce a novel prompting technique optimized for semi-structured documents. \cite{zspot} developed zero-shot prompt templates for question answering over financial documents that guide the LLMs to encode reasoning into a python program. These efforts do not consider fine-tuning to specialize language models for the financial domain. \cite{llm_tool} explore task-wise integration of external tools, calculator and SQL engine, with fine-tuned small language models for tabular data analysis in finance. However, their approach utilizes predetermined prompt templates for data generation rather than a teacher-student approach for fine-tuning. In our approach, we generate structured python programs using pre-trained LLMs and train small language models with supervised fine-tuning. Furthermore, we examine in detail the nature of the alignment achieved by the fine-tuning for the financial question answering, an area that is largely unexplored in previous research.

\section{Fine-tuning for Financial Question Answering} \label{sec:fine-tuning}

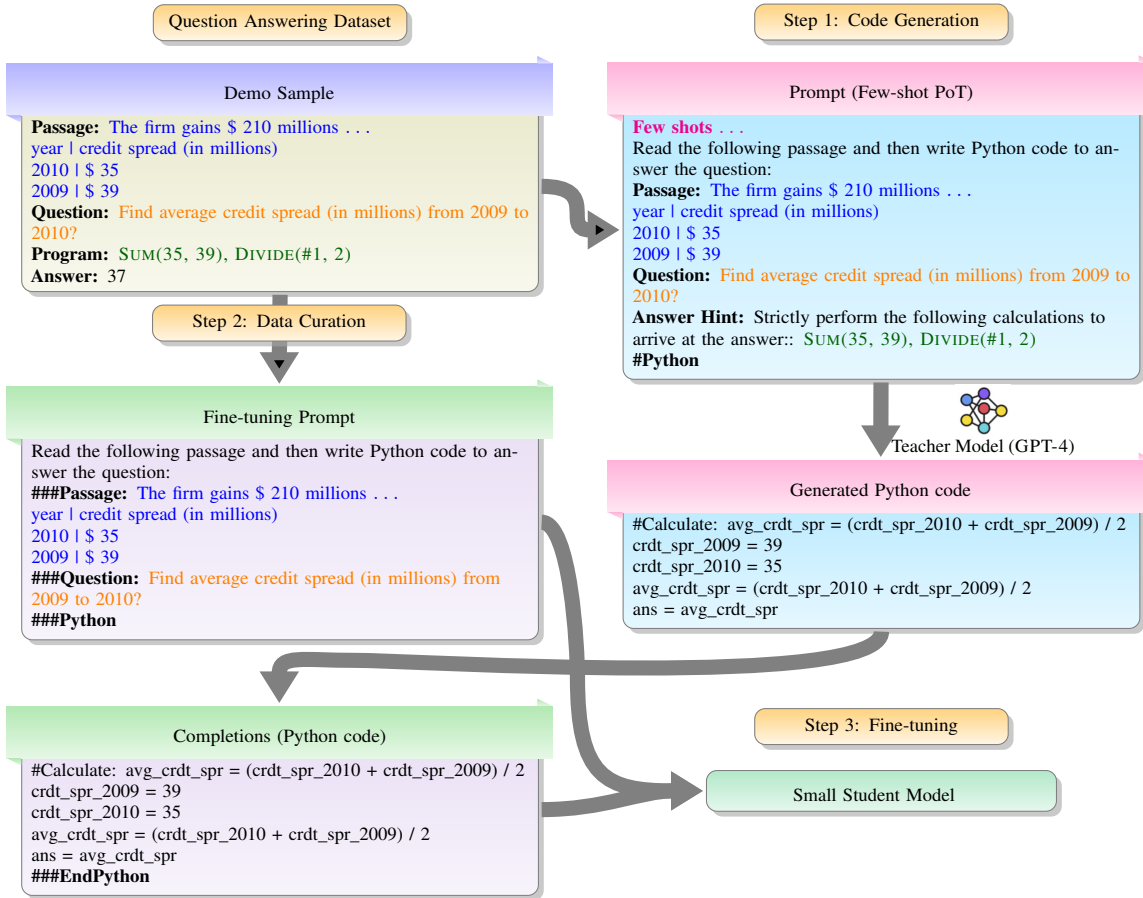
\begin{figure*}[htb]
\begin{center}
\begin{tikzpicture}[node distance=11mm]%
	\node (m) [miniprompt=cyan] {\\\phantom{hello} \\\clm{\textbf{Few shots $\ldots$}} \\Read the following passage and then write Python code to answer the question:\\ 
	\textbf{Passage:} \clb{The firm gains \$ 210 millions $\ldots$\\
year | credit spread (in millions)\\
2010 | \$ 35\\
2009 | \$ 39\\}
	\textbf{Question:} \clo{Find average credit spread (in millions) from 2009 to 2010?}\\
	\textbf{Answer Hint:} Strictly perform the following calculations to arrive at the answer:: \cldg{\scr{Sum(35, 39), Divide(\#1, 2)}}\\
	\textbf{\#Python}
}; 
	\addHeader{m}{mh}{magenta}{red}{Prompt (Few-shot PoT)}

	\node (n) [miniprompt=yellow, left=of m, yshift=1.5em] {\\\phantom{hello} \\ 
	\textbf{Passage:} \clb{The firm gains \$ 210 millions $\ldots$\\
year | credit spread (in millions)\\
2010 | \$ 35\\
2009 | \$ 39\\}
	\textbf{Question:} \clo{Find average credit spread (in millions) from 2009 to 2010?}\\
	\textbf{Program:} \cldg{\scr{Sum(35, 39), Divide(\#1, 2)}}\\ 
	\textbf{Answer:} 37   
}; 
	\addHeader{n}{nh}{blue}{blue}{Demo Sample}

	\node (p) [miniprompt=lightred, below=of n, yshift=-1em] {\\\phantom{hello} \\Read the following passage and then write Python code to answer the question:\\ 
	\textbf{\#\#\#Passage:} \clb{The firm gains \$ 210 millions $\ldots$\\
year | credit spread (in millions)\\
2010 | \$ 35\\
2009 | \$ 39\\}
	\textbf{\#\#\#Question:} \clo{Find average credit spread (in millions) from 2009 to 2010?}\\
	\textbf{\#\#\#Python}
}; 
	\addHeader{p}{ph}{darkgreen}{green}{Fine-tuning Prompt}

	\node (c) [miniprompt=lightred, below=of p] {\\\phantom{hello} \\ 
\#Calculate: avg\_crdt\_spr = (crdt\_spr\_2010 + crdt\_spr\_2009) / 2\\
crdt\_spr\_2009 = 39\\
crdt\_spr\_2010 = 35\\
avg\_crdt\_spr = (crdt\_spr\_2010 + crdt\_spr\_2009) / 2\\
ans = avg\_crdt\_spr\\
\textbf{\#\#\#EndPython}
}; 
	\addHeader{c}{ch}{darkgreen}{green}{Completions (Python code)}

	\node (d) [miniprompt=cyan, below=of m, yshift=-0.5em] {\\\phantom{hello} \\ 
\#Calculate: avg\_crdt\_spr = (crdt\_spr\_2010 + crdt\_spr\_2009) / 2\\
crdt\_spr\_2009 = 39\\
crdt\_spr\_2010 = 35\\
avg\_crdt\_spr = (crdt\_spr\_2010 + crdt\_spr\_2009) / 2\\
ans = avg\_crdt\_spr
}; 
	\addHeader{d}{dh}{magenta}{red}{Generated Python code}

	\node (e) [ultraminiprompt=green, below=of d, yshift=-2em] {\\\phantom{Some sm} Small Student Model};

	\draw [arrow] (n) edge[out=0, in=180, looseness=1.7] (m);
	\draw [arrow] (n) edge[out=270, in=90, looseness=0.5] (ph);
	\draw [arrow] (c) edge[out=0, in=180, looseness=1] (e);
	\draw [arrow] (p) to[out=0, in=90, looseness=0.5] ($(e)+(-4,1.5)$) to[out=270, in=180, looseness=1.5] (e);
	\draw [arrow] (d) to[out=270, in=0, looseness=0.3] ($(ch)+(1,0.95)$) to[out=180, in=90, looseness=1.5] (ch);
	\draw [arrow] (m) --  (dh) node[midway, above] {} coordinate (amdh);
	\node[above right= of amdh, xshift=0.2em, yshift=-0.7em] (image) {\includegraphics[height=0.7cm]{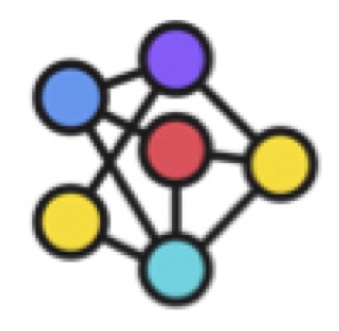}};
	\node[below= -0.25cm of image] {\scriptsize Teacher Model (\gpt)};

	\node (headm) [heading, above=of m, yshift=-1.5em] {Step 1: Code Generation};
	\node (headn) [heading, above=of n, yshift=-1.5em] {Question Answering Dataset};
	\node (heade) [heading, above=of e, yshift=-2em] {Step 3: Fine-tuning};
	\node (headp) [heading, above=of ph, yshift=-1.3em] {Step 2: Data Curation};
\end{tikzpicture}

\end{center}
\caption{Fine-tuning for financial question answering.}
\label{fig:fine-tuning}
\end{figure*}

We adopt the approach of using very large-scale models as reasoning teachers, and fine-tuning relatively small-scale student models from the prompt-completion pairs generated using the teacher model \cite{distill_step_by_step,reasoning_teachers}. Specifically, the large model is used to generate python code with comments that encapsulates the reasoning required to answer the financial question. Furthermore, the programs are generated with a specific structure that facilitates subsequent performance analysis of the fine-tuned model, as discussed in detail in Sec. \ref{sec:performance_analysis}. The fine-tuning task is performed in three steps as shown in Figure \ref{fig:fine-tuning}.

\subsection{Code Generation}
\begin{figure}[htb]
\begin{center}
\begin{tikzpicture}
	\node (n1) [inline_prompt] {
\clm{\#Calculate: avg\_crdt\_spr = (crdt\_spr\_2010 + crdt\_spr\_2009) / 2\\}
\cldg{crdt\_spr\_2009 = 39\\
crdt\_spr\_2010 = 35\\}
{\color{violet}avg\_crdt\_spr = (crdt\_spr\_2010 + crdt\_spr\_2009) / 2\\
ans = avg\_crdt\_spr}
}; 
\node [fill=magenta, minimum width=0.05cm, minimum height=0.05cm, node distance=3mm, below left=0.25cm and -0.55cm of n1](c1) {};
\node [right= -0.05cm of c1](cn1) {\scriptsize \clm{Concept}};
\node [fill=darkgreen, minimum width=0.05cm, minimum height=0.05cm, right=1.5em of cn1](c2) {};
\node [right= -0.05cm of c2](cn2) {\scriptsize \cldg{Extracted Entities}};
\node [fill=violet, minimum width=0.05cm, minimum height=0.05cm, right= 1.5em of cn2](c3) {};
\node [right= -0.05cm of c3](cn3) {\scriptsize {\color{violet}Remaining Code}};
\end{tikzpicture}
\end{center}
\caption{\gpt generated sample Python code}
\label{fig:code_seg}
\end{figure}
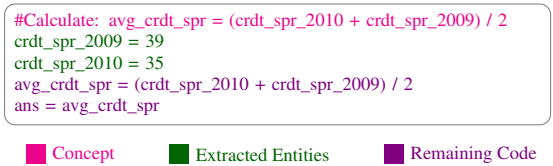

In the code generation step, we employ the teacher model to generate a Python code for a specified question-answering task. Financial question answering consists of three distinct steps: understanding the concept and writing the formula required to answer the question, finding the relevant entities, and then executing and storing the calculations. A sample of the desired code structure encapsulating this reasoning process is show in in Figure \ref{fig:code_seg}. We guide the teacher model to consistently generate the desired code structure through program of thought (PoT) prompting \cite{pot}. In few-shot PoT prompting, as shown in Figure \ref{fig:fine-tuning}, few shot exemplars are prefixed as demonstrations for the teacher model to generate codes in the desired format. 

In the financial question answering training dataset, each sample contains the final answer to the question and additionally it may contain a program which demonstrate step-wise arithmetic calculations that are required to be performed to arrive at the answer. We incorporate these programs as answer hints in the few-shot PoT prompt to guide the teacher model towards accurate code generation. This strategy can potentially improve the question-answering accuracy of the teacher model.  

\subsection{Data Curation}
During data curation, we filter out the samples with incorrect teacher codes and format the filtered samples to prompt-completion pairs for the student model. At the filtering stage, the samples with incorrect teacher-generated codes are identified by executing the codes and comparing the resulting answer with the ground truth answer. The filtered samples are then formatted to prompt-completion pairs as per student model requirements. For instance, the prompt instructions for the \Mistral 7B (instruction tuned) model should begin with the token [INST] and ends with the token [/INST]. In addition, special characters like `\#' can be used to symbolize the prompt structure.          

\subsection{Fine-tuning}
The fine-tuning task for question answering is represented by the prompt-completion pairs: $\mathcal{D}=\{x_i, y_i\}_{i=1}^{N}$ where $x_i$ is a token sequence for the fine-tuning prompt and $y_i$ is a token sequence for the corresponding code completion, as shown in Figure \ref{fig:fine-tuning}. We use low rank adaptation (LoRA) \cite{lora}, a special class of parameter efficient fine-tuning that takes advantage of the low ``intrinsic dimension'' of pre-trained LLMs, when adapting to a specific task \cite{intrinsic}. The student model's LoRA adapter is fine-tuned to adapt to the financial question answering task.

\section{Experiments}
\subsection{Experimental Design}
\phantom{us}\textbf{Datasets:}
We conduct our experiments on three English language financial question answering datasets FinQA \cite{finqa}, ConvFinQA \cite{convfinqa} and TATQA \cite{tatqa}. The question answering task, in FinQA and TATQA datasets, is to answer the questions using the passage containing text and table content. The experiments for TATQA are restricted to questions of \textit{arithmetic} type. In ConvFinQA, the task is to answer the last question from a conversation containing a series of interrelated questions, based on a given text and table content.  The table content is represented in a textual format using the strategy adopted in \cite{chen}.  

\textbf{Fine-tuning:}
In our experiments, for the teacher model we use the \gpt model provided by Azure OpenAI\footnote{https://oai.azure.com/}. For student models, we consider the following open source family of models: \fPhi, \Mistral and \Orca. We provide further details on the models in Table \ref{tab:llms} in the Appendix. In the code generation step, we use \gpt as a teacher, and prompt it with 4-shot exemplars for all datasets. These exemplars are derived from few-shot PoT prompts as discussed in \cite{pot}.  The few-shot teacher's prompt for FinQA, ConvFinQA and TATQA dataset is given in Figure \ref{fig:few_shot_finqa}, Figure \ref{fig:few_shot_convfinqa} and Figure \ref{fig:few_shot_tatqa} in the Appendix \ref{app:few_shots} respectively. In each of the datasets, an expression encoding the required calculations, is provided for each training sample. We include the expression as a hint in the prompt for the \gpt model. In the data curation step, any data that contains incorrect \gpt code is eliminated. 
The filtered data is then converted into prompt-completion pairs to meet the requirements of the student model. Finally, during the fine-tuning stage, the student model is trained using these prompt-completion pairs. For fine-tuning the student models,  we use LoRA \cite{lora} method of  HuggingFace PEFT library\footnote{https://huggingface.co/docs/peft/en/index} that backpropagates gradients to low rank adapters.

In our fine-tuning experiments, the datasets are divided into train, dev, and test splits. The FinQA dataset includes predefined splits with their ground truths. As the test splits for ConvFinQA and TATQA datasets don't include ground truths, we use the predefined dev splits of these datasets as the test splits. The dev splits for ConvFinQA and TATQA are created by randomly picking 10\% samples of their predefined train splits and the remaining samples are used for the training. The datasets with their splits are summarized in Table \ref{tab:datasets}. 
\begin{table}[htb]
\centering
\begin{threeparttable}
\begin{tabular}{llll}
\hline
	\textbf{Dataset} & \textbf{Train} & \textbf{Dev} & \textbf{Test}\\
\hline
	FinQA & 6251 &  883 & 1147 \\
\hline
	ConvFinQA & 2737 &  300 & 421 \\
\hline
	TATQA\tnote{$\dagger$} & 4992 &  550 & 718 \\
\hline
\end{tabular}
\begin{tablenotes}
\item[$\dagger$] {\footnotesize Only arithmetic questions are considered.}
\end{tablenotes}
\end{threeparttable}
\caption{\label{tab:datasets}Datasets for fine-tuning}
\end{table}
The train and dev splits are used for the student model's fine-tuning and validation respectively. We select the model and the hyperparameters\footnote{The set of explored and optimal hyperparameters are provided in Table \ref{tab:hyperparameters} in the Appendix \ref{app:hyperparameters}.} that give the highest performance on the dev split, and report the fine-tuned model's performance on the test split. We employ the vLLM\footnote{https://github.com/vllm-project/vllm} framework for conducting inference on fine-tuned models. The experiments are performed on a compute instance with 24 cores, 220GB RAM and a A100 GPU (80GB). 

\textbf{Evaluation Metrics:} 
The fine-tuned LLMs are trained to generate Python codes that are executed using the Python \texttt{exec} function to determine the resulting answer. The resulting answer is then compared against the ground truth, using the method described in \cite{zspot}.

\textbf{Baselines:} 
We evaluate the base version of the student models using zero-shot and few-shot prompting.

\textit{Zero-shot prompting:} We performed prompt engineering experiments for synthesizing the zero-shot prompt that directs the LLMs to generate Python code to answer the question. For the FinQA dataset, the optimized zero-shot prompt for the \fPhi, \Orca and \Mistral models is given in Figure \ref{fig:zero_shot_prompts} in the Appendix.

\textit{Few-shot prompting:} The few-shot prompt includes a few example demonstrations for LLMs to learn in-context. For the FinQA dataset, the few-shot prompt for \fPhi, \Orca and \Mistral models is given in Figure \ref{fig:few_shot_prompts} in the Appendix. 

\subsection{Fine-tuning Results}
The performance of the fine-tuned LLMs is reported in Table \ref{tab:fine-tuning-accuracy}. For comparison, we report the performance of zero-shot and few-shot prompts using \cgpt and \gpt. These results are taken from \cite{zspot}. The zero-shot prompt methods are defined as ZS-FinPYT (\cgpt) and ZS-FinPYT (\gpt) while the few-shot methods are defined as Few-shot PoT (\cgpt) and Few-shot PoT (\gpt). In the results discussed below, the comparison of the fine-tuned models is done against the best-performing prompting method for a given dataset and GPT model. For conciseness we refer to only the model’s name instead of the full method.

\textbf{FinQA dataset:}
The zero-shot and few-shot performance of the models is highly varying with some models having low zero-shot accuracy while other achieve excellent zero shot performance, with one of the models surpassing \cgpt. For some models, the few-shot results show significant improvement as compared to their zero-shot performance while for others the few-shot prompting results in a lower accuracy. The fine-tuned models show large improvements over their respective zero-shot and few-shot results. Despite major differences in performance of the base models, the fine-tuned versions achieve a similar accuracy within a 6\% range. The results thus demonstrate that the proposed fine-tuning approach was effective across models with a wide range of performance for the base models. For fine-tuned models, the model size has a small effect with the larger \Orca model being 2\% more accurate than the smaller one while \Phimd has 4\% higher accuracy as compared to \Phimn. All the fine-tuned models outperform \cgpt by 4\%-10\%, with the \Phimd achieving an accuracy within 1\% of the \gpt. Overall, the results indicate that relatively small language models can be fine-tuned to be competitive with much larger models, for the financial question answering task. 

\textbf{ConvFinQA dataset:}
The results broadly follow a similar pattern to the results of FinQA but there are some differences. The differences between the fine-tuned models are slightly pronounced as compared to the FinQA dataset. The models are within 10\% of each other with the effect of model size being more significant. For both \Orca and \fPhi, the larger models achieve $\sim$5\% higher accuracy than their respective smaller variants. 

\textbf{TATQA dataset:}
The results for TATQA closely mirror those observed for the FinQA dataset with all the models achieving excellent performance, their accuracies falling within a 5\% range. The fine-tuned \Phimd model excels for this dataset, marginally surpassing \gpt. Model size has minimal effect for \Orca while for \fPhi a small effect is observed, with \Phimd achieving 3\% higher accuracy than \Phimn. 

\begin{table*}[htb]
\setlength{\tabcolsep}{4.5pt}
\centering
\begin{threeparttable}
\begin{tabular}{lccccccc}
\hline
\textbf{Methods} & \textbf{\sMistral-7B}\tnote{$\star$} & \textbf{\sOrca-7B}\tnote{$\star$} & \textbf{\sOrca-13B}\tnote{$\star$} & \textbf{\sPhi-3.8B}\tnote{$\star$} & \textbf{\sPhi-14B}\tnote{$\star$} & \textbf{\scgpt} & \textbf{\gpt}\\
\hline
\multicolumn{8}{l}{\textbf{FinQA Accuracy}}\\
\hline
zero-shot & 41.58 & 2.78 & 37.49 & 57.62 & 70.35 & 66.52 & \textbf{77.51}\\ 
\hline
few-shot\tnote{$\dagger$} & 50.82 & 15.34 & 4.36 & 42.45 & 66.95 & 67.39 & \textbf{78.46}\\
\hline
fine-tuned & 76.63 & 71.57 & 73.75 & 73.49 & \textbf{77.59} & - & -\\
\hline
\multicolumn{8}{l}{\textbf{ConvFinQA Accuracy}}\\
\hline
zero-shot & 31.82 & 2.61 & 28.03 & 48.21 & 59.38 & 67.45 & \textbf{76.95}\\ 
\hline
few-shot\tnote{$\dagger$} & 18.52 & 10.68 & 11.88 & 41.33 & 60.09 & 65.79 & \textbf{82.42}\\
\hline
fine-tuned & 76.48 & 70.30 & 75.77 & 76.00 & \textbf{81.94} & - & -\\ 
\hline
\multicolumn{8}{l}{\textbf{TATQA Accuracy}}\\
\hline
zero-shot & 66.01 & 4.59 & 44.29 & 78.27 & 85.37 & 85.00 & \textbf{93.00}\\ 
\hline
few-shot\tnote{$\dagger$} & 66.85 & 26.88 & 17.83 & 59.33 & 76.32 & 74.75 & \textbf{91.89}\\
\hline
fine-tuned & 88.71 & 88.57 & 88.86 & 90.52 & \textbf{93.73} & - & -\\ 
\hline
\end{tabular}
\begin{tablenotes}
      \item[$\star$]{\footnotesize \sMistral-7B, \sOrca-\{\texttt{x}\}B, \sPhi-\{\texttt{x}\}B, and \scgpt represent \Mistral 7B, \Orca-\{\texttt{x}\}B, \fPhi-\{\texttt{x}\}B, and \cgpt respectively.}
      \item[$\dagger$] {\footnotesize Few-shot PoT prompting is used with 4-shots selected from the few-shots used in \cite{pot}.}
\end{tablenotes}
\end{threeparttable}
\caption{\label{tab:fine-tuning-accuracy}
Accuracy of the fine-tuned models on financial datasets.
}
\end{table*}

\section{Performance Analysis}\label{sec:performance_analysis}
We examine the evolution of model capabilities during fine-tuning, seeking to identify specific model enhancements. To this end, we define, and measure three key capabilities (1) concept understanding measured by the ability to correctly identify the required calculation (2) entity extraction measured by the ability to extract all required entities and (3) generation of executable code. Our method relies on comparing the output of the student models with that of the teacher generated codes. Therefore, we remove all samples with incorrect teacher generated codes from further analysis. 

\begin{table*}[htb]
\centering
\begin{threeparttable}
\begin{tabular}{lccccccc}
\hline
\multirow{2}{*}{\textbf{Fine-tuned Models}} & \multicolumn{2}{c}{\textbf{Concept}}& \multicolumn{2}{c}{\textbf{Entity Extraction}}& \multicolumn{2}{c}{\textbf{Executable Code}}\\
\cline{2-7}
 & Base model & Epoch-1 & Base model & Epoch-1 & Base model & Epoch-1\\
\hline
\textbf{\Mistral 7B} & 53.09 & 84.18 & 66.99 & 90.72 & 56.73 & 99.54\\  
\hline
\textbf{\Orca-7B} & 27.94 & 76.89 & 57.54 & 90.71 & 10.87 & 91.95\\ 
\hline
\textbf{\Orca-13B} & 46.48 & 79.6 & 71.78 & 92.69 & 29.67 & 91.95\\
\hline
\textbf{\Phimn} & 69.22 & 82.94 & 82.81 & 90.23 & 94.33 & 96.82\\ 
\hline
\textbf{\Phimd} & 79.23 & 84.55 & 65.26 & 90.48 & 98.18 & 96.94\\
\hline
\end{tabular}
\end{threeparttable}
\caption{\label{tab:concept_and_entity_accuracy}
Evaluation of concept, entity extraction and executable code accuracy by \gpt for the FinQA dataset 
}
\end{table*}

\subsection{Concept accuracy}
Due to significant model output variation, it is hard to define a simple metric to measure concept accuracy. Motivated by the promising results shown by others \cite{llm_judge} in using LLM for evaluation in challenging scenarios, we propose the use of GPT-4 to rate the concept understanding demonstrated in the model output, using a 5-point scale defined as follows: 1: no understanding 2: limited understanding 3: partial understanding 4: mostly demonstrates understanding 5: perfect understanding. After some prompt engineering using 25 random student code samples, we found this method to provide reasonable assessment of concept accuracy. The key instructions needed were to guide the evaluator to focus on the presence of relevant entities, and not their values or the output format. The final prompt which includes the instructions, the output of the student model, the gold code and the question are shown in Figure \ref{fig:gpt_judge_concept} in the Appendix. 

We define the concept accuracy as the percentage of cases where the student model output receives a rating of 5. For all models, we measure the concept accuracy for the base model as well as checkpoint after one epoch (see Figure \ref{fig:gpt_rating_orca7b}--\ref{fig:gpt_rating_phimd} in the Appendix for rating distributions). The concept accuracy shown in Table \ref{tab:concept_and_entity_accuracy}, indicates the \Orca family models have significantly lower concept accuracy initially as compared to other models. However, the fine-tuning leads to substantial improvements in these models, leading to a small gap in concept accuracy as compared to other models post fine-tuning. To better understand these results, we manually examined fifty random sample outputs from each of the base models, that were assessed as lacking concept accuracy. We then examined the output of the models for the same samples after one epoch. While the analysis was performed on all models, in the following sections we present the detailed analysis of \Phimn and \Orca-7B models (See Appendix \ref{app:examples} for many illustrative examples). The analysis of the remaining models reveals similar patterns.

\textbf{\Phimn concept accuracy:}
For the base \Phimn model, the overall concept accuracy was around 70\% with about 16\% samples receiving a rating of 1 or 2 by \gpt. The \Phimn model responses don’t follow a standard structure while  answering the question. In a significant number of samples with low concept rating, the base model's response does not provide the formulas/arithmetic steps that are required to answer the question, thus failing to demonstrate concept understanding. About 7\% of the samples received a rating of 4, with many of these responses containing minor arithmetic errors, providing formulas with closely related but not correct entities, and other small issues.
Most of the base model responses with missing formulas are corrected by training the model for one epoch. Approximately 80\% of cases, initially rated 4, are also corrected after one epoch. Several cases where the formula was properly but incorrectly written by the base model remained incorrect even after 1 epoch. 

\textbf{\Orca-7B concept accuracy:}
The initial \Orca model provides a long explanation of the required reasoning, often failing to produce executable code. Sometimes the formulae are written descriptively without use of mathematical representations. In a significant number of cases, the model includes the input passage leading to an incomplete output that does not contain the required formula. As a result, 43\% of the samples received a rating of 1 or 2 and the model has a low concept accuracy of 28\%. Occasionally, the long explanations do provide the correct formula which is not identified by the \gpt. These results are likely due to the \Orca being a model that is specialized to solve reasoning problems using elaborate reasoning traces and not being explicitly trained in generating code that encodes the reasoning. Hence the \Orca  model tends to produce output that is significantly different from that expected for the specific problem formulation considered in this study.

Upon fine-tuning, the model quickly learns to produce executable code with the formula written out in the desired format, while suppressing unnecessary output. We observe that the model does correct some of the formula mistakes that it made initially. Similar to \Phimn, we observed that many cases (12\% of the samples) initially had a 4 rating, triggered by minor errors. A vast majority of these cases get corrected during fine-tuning. These improvements are reflected in the fine-tuned \Orca model achieving a significantly improved concept accuracy of 77\% after one epoch.

\begin{table*}[htb]
\centering
\begin{threeparttable}
\begin{tabular}{lcccc}
\hline
\multirow{3}{*}{\textbf{Fine-tuned Models}} & \multicolumn{4}{c}{\textbf{Training dataset} (samples from train split)}\\
\cline{2-5}
 & \multirow{2}{*}{FinQA:5698}  & \multirow{2}{*}{FinQA:1500}  & FinQA:1000 & \multirow{2}{*}{ConvFinQA:2550}\\
& & & ConvFinQA: 500 & \\ 
\hline
\multicolumn{5}{l}{\textbf{FinQA accuracy}}\\
\hline
\Mistral 7B & 76.63  & 70.35  & 68.78  & 64.95 \\  
\hline
\Orca-7B & 71.57 & 63.56 & 67.13 & 55.01 \\ 
\hline
\Orca-13B & 73.75  & 70.53  & 70.09 & 65.13 \\
\hline
\Phimn & 73.49  & 69.83  & 71.14  & 68.87 \\ 
\hline
\Phimd & 77.59 & 74.71 & 75.06 & 74.80 \\
\hline
\multicolumn{5}{l}{\textbf{ConvFinQA accuracy}}\\
\hline
\Mistral 7B & 36.34  & 33.01  & 72.20  & 76.48 \\  
\hline
\Orca-7B & 41.09  & 32.3  & 69.12  & 70.30 \\ 
\hline
\Orca-13B & 36.82  & 41.81  & 72.92  &  75.77 \\
\hline
\Phimn & 44.89  & 49.40  & 72.20  & 76.00 \\ 
\hline
\Phimd & 53.44 & 51.54 & 79.57 & 81.94 \\
\hline
\end{tabular}
\end{threeparttable}
\caption{\label{tab:small_dataset_accuracy}
Effect of training data size on fine-tuning.  
}
\end{table*}

\subsection{Entity extraction}
To measure the entity extraction capability of the student models, we incorporate the first line of the teacher model’s code into the fine-tuning prompt and perform inference on the fine-tuned model to complete the Python code. Since the provided line is a comment with the formula required to answer the question, the main task is to extract the required entities. We then use \gpt with the prompt shown in Figure \ref{fig:gpt_judge_entity} in the Appendix, to assess the student code and determine if all the required entities have been correctly extracted.

The results shown in Table \ref{tab:concept_and_entity_accuracy} indicate significant improvement in entity extraction capability of all the student models during the fine- tuning, with all of them showing similar accuracy after fine-tuning. The results suggest that the fine-tuning helps the base model adapt to the specific table structures and data format present in the financial dataset, improving the entity extraction performance and contributing to overall model accuracy.

\subsection{Code generation}
As the required code for this problem is simple, we only use the ability to generate executable code as a measure of accuracy. The results are shown in Table \ref{tab:concept_and_entity_accuracy}. As compared to other models, the base \fPhi models have very high percentage of cases where they generate executable code. After epoch 1, all models have more than 90\% success rate of producing executable code, indicating that the models initially lacking the ability to generate the required code, have significantly improved.

Overall, the detailed assessment revealed two main reasons that leads to improved model performance post fine-tuning: (1) Training with a standard reasoning chain induces the models to express and apply the required concept and (2) Adapting to the specific data format improves the entity extraction performance.

\subsection{Effect of training data size}
Given the observed effects of fine-tuning, it naturally leads us to inquire whether a smaller dataset might be adequate for achieving similar enhancements. Another related and interesting question is the volume of data necessary to adapt the FinQA model for proficiency with ConvFinQA, which, while originating from the same domain as FinQA, introduces the additional complexity of processing conversational-style questions.
 
We perform fine-tuning experiments with the following settings to understand the data requirements (a) 1500 data points randomly sampled from the original FinQA dataset and (b) 1000 samples randomly sampled from the FinQA dataset combined with 500 samples randomly sampled from ConvFinQA dataset. The models fine-tuned on these datasets is compared with the corresponding models trained on the entire FinQA and ConvFinQA training datasets, respectively. Evaluation on the test data of both datasets is shown in Table \ref{tab:small_dataset_accuracy}.
 
The model trained on 1500 data points from FinQA is within 3\%-8\% of the model trained with the entire data, suggesting that effective fine-tuning can be achieved with a significantly smaller dataset. The models trained on the entire FinQA dataset perform poorly when directly used on the ConvFinQA test data. However, when the models are trained with a smaller dataset consisting of 1000 FinQA samples and 500 samples from ConvFinQA data, there is improvement of more than 25\% across models, with all models achieving an accuracy within 5\% of the model trained with the entire ConvFinQA training data.  These findings suggest that the financial concept understanding along with fine-tuning provides most of the key learning that can address ConvFinQA questions as well. However, the model needs a small sample from the ConvFinQA dataset to adapt to the conversational style of questions used in that dataset.

\section{Conclusion}
We explored the performance of small language models fine-tuned for financial question answering, using exemplars generated by a very large teacher model.  The small models achieved accuracy comparable to that of the teacher model, driven primarily by improved ability to apply financial concepts as well as entity extraction. We showed that smaller datasets can yield similar results, suggesting that small language models can be efficiently fine-tuned for complex financial domain problems.

\section*{Limitations}
The use of \gpt to assess concept understanding using the base and fine-tuned student models’ output, can sometimes produce erroneous determinations of concept error. While we instruct \gpt to not assess based on the output format, we found that elaborate responses could sometimes lead to a false assessment. Despite this limitation, the method is still effective in achieving our primary goal of understanding the effect of fine-tuning for financial question answering.

While we perform hyperparameter optimization for fine-tuning the student models, the small differences between the performance of the fine-tuned models could be simply due the hyperparameters not being fully optimal. Since we focus more on the improvements achieved in the fine-tuned model over their corresponding base model,  this doesn’t have a major impact on the findings reported in the paper.

Our experiments are limited to only on the single task of financial question answering. The performance and behaviour of the small models in a multi-task setup needs to be explored in the future.

While we demonstrate that small language models can achieve performance approaching that of much larger models, they also inherit some of the associated risks. For cases where the reasoning was incorrect, the current system will provide an explanation with a high level of confidence, which can be misleading. Our models currently does not address or control for such behavior and we have not studied the nature or extent of this problem. In practice, this can pose challenges for practical use in real world systems. For real world use, a human would need to review the output prior to using any of the information generated. Future research to understand this potential risk in more detail and provide indications of when the model is not sure of its response would be valuable. 

\section*{Disclaimer}
The views reflected in this article are the views of the authors and do not necessarily reflect the views of the global EY organization or its member firms.

\bibliography{references}
\bibliographystyle{acl_natbib}

\appendix

\section{Fine-tuning: Experimental Details}\label{app:hyperparameters}
The student models are fine-tuned using LoRA module of HuggingFace's PEFT library. We experimented with learning rates of $2.5\text{e}^{-5}$, $5\text{e}^{-5}$, $1\text{e}^{-4}$ and LoRA $r$ parameters 64, 256, 512, 1024 for all the models. The optimal hyperparameters used for fine-tuning are outlined in Table \ref{tab:hyperparameters} and are categorized into three parts: 
\begin{enumerate}[leftmargin=*]
\item \textbf{LoRA parameters}, which comprise of LoRA config parameters of the HuggingFace's PEFT library\footnote{https://github.com/huggingface/peft}.     
\item \textbf{Training parameters} include config settings of SFTTrainer of the HuggingFace's  TRL library\footnote{https://huggingface.co/docs/trl/v0.7.2/en/sft\_trainer}.
\item \textbf{Inference parameters} are associated with the vLLM\footnote{https://github.com/vllm-project/vllm} settings needed for the inference of fine-tuned models.   
\end{enumerate}

\begin{table}
\centering
\begin{threeparttable}
\begin{tabular}{ll}
\hline
\multicolumn{2}{l}{\textbf{LoRA parameters}}\\
\hline
LoRA\_{$r$} & 512 \\
LoRA\_{$\alpha$} & 1024 \\
LoRA\_{\text{dropout}} & 0.1 \\
LoRA\_{\text{bias}} & \text{none}\\
PEFT task\_type \phantom{cc} & CAUSAL\_LM \\
target\_modules & all linear \\ 
\hline
\multicolumn{2}{l}{\textbf{Training parameters}}\\
\hline
epoch & 6 \\
batch size & 1\\
gradient accum steps & 1\\
learning rate (\Mistral) & $2.5\text{e}^{-5}$ \\
learning rate (\Orca) & $5\text{e}^{-5}$ \\
learning rate (\fPhi) & $1\text{e}^{-4}$ \\
bf16 & True \\ 
\hline
\multicolumn{2}{l}{\textbf{Inference parameters (vLLM)}}\\
\hline
temperature & 0\\
top\_p & 0.9 \\
max\_tokens & 1000\\
\hline
\end{tabular}
\end{threeparttable}
\caption{\label{tab:hyperparameters}Hyperparameters used for fine-tuning}
\end{table}

\begin{table*}
\centering
\begin{threeparttable}
\begin{tabular}{llll}
\hline
\textbf{Model Name} & \textbf{Parameters} & \textbf{HuggingFace API} & \textbf{License}\\
\hline
\Mistral 7B & 7B & \texttt{mistralai/Mistral-7B-Instruct-v0.2} & apache-2.0\\
\Orca-7B & 7B & \texttt{microsoft/Orca-2-7b}  & msr-lic\tnote{$\star$}\\
\Orca-13B & 13B & \texttt{microsoft/Orca-2-13b} & msr-lic\tnote{$\star$}\\
\Phimn & 3.8B & \texttt{microsoft/Phi-3-mini-128k-instruct} & mit \\
\Phimd & 14B & \texttt{microsoft/Phi-3-medium-128k-instruct} & mit \\
\hline
\end{tabular}
\begin{tablenotes}
      \item[$\star$]{msr-lic stands for microsoft-research-license}
\end{tablenotes}
\end{threeparttable}
\caption{\label{tab:llms}Description of student LLMs used for fine-tuning}
\end{table*}
The additional details on the student models including license and terms of use is provided in Table \ref{tab:llms}. We use the \Orca models in a manner consistent with their intended use:
 
As described in the website: \url{https://www.microsoft.com/en-us/research/publication/orca-2-teaching-small-language-models-how-to-reason/},
\Orca models were designed for research settings and should be used only for research purposes, and its testing has only been carried out in such environments. It should not be used in downstream applications.
 
We further fine-tune \Orca and use it only for research purposes, and will not use it for any commercial purposes. 

\section{Fine-tuning and Baseline Prompts}\label{app:prompts}
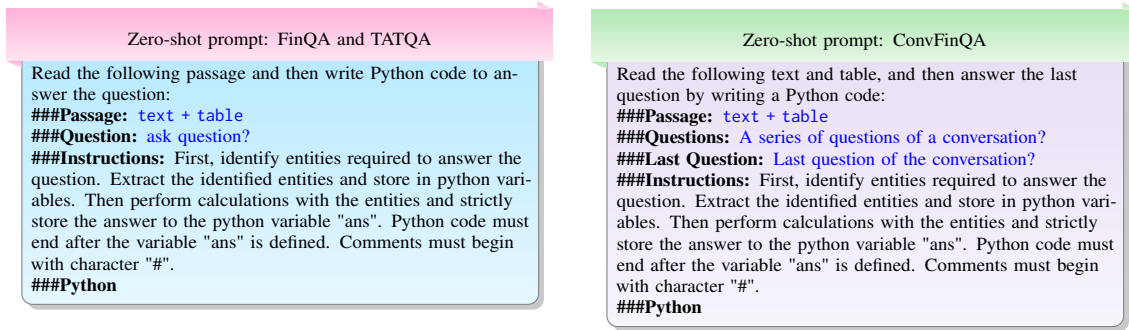
\begin{figure*}
\begin{center}
\begin{tikzpicture}[node distance=5mm]
\node (p) [miniprompt=cyan, yshift=-1em] {\\\phantom{hello} \\Read the following passage and then write Python code to answer the question:\\
        \textbf{\#\#\#Passage:} \clb{\texttt{text + table}}\\
        \textbf{\#\#\#Question:} \clb{ask question?}\\
	\textbf{\#\#\#Instructions:} First, identify entities required to answer the
question. Extract the identified entities and store in python variables.
Then perform calculations with the entities and strictly
store the answer to the python variable "ans". Python code must
end after the variable "ans" is defined. Comments must begin
with character "\#".\\
        \textbf{\#\#\#Python}\\
};
        \addHeader{p}{ph}{magenta}{red}{Zero-shot prompt: FinQA and TATQA}
\node (q) [miniprompt=lightred,right=of m, yshift=-1.4em, xshift=1em] {\\\phantom{hello} \\Read the following text and table, and then answer the last question by writing a Python code:\\
        \textbf{\#\#\#Passage:} \clb{\texttt{text + table}}\\
        \textbf{\#\#\#Questions:} \clb{A series of questions of a conversation?}\\
        \textbf{\#\#\#Last Question:} \clb{Last question of the conversation?}\\
	\textbf{\#\#\#Instructions:} First, identify entities required to answer the
question. Extract the identified entities and store in python variables.
Then perform calculations with the entities and strictly
store the answer to the python variable "ans". Python code must
end after the variable "ans" is defined. Comments must begin
with character "\#".\\
        \textbf{\#\#\#Python}\\
};
        \addHeader{q}{qh}{darkgreen}{green}{Zero-shot prompt: ConvFinQA}
\end{tikzpicture}
\end{center}
\caption{Zero-shot prompts. \label{fig:zero_shot_prompts}}
\end{figure*}

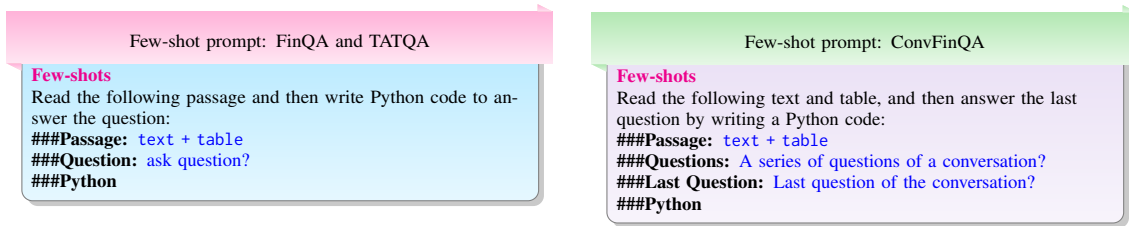
\begin{figure*}
\begin{center}
\begin{tikzpicture}[node distance=5mm]
\node (p) [miniprompt=cyan, yshift=-1em] {\\\phantom{hello} \\\clm{\textbf{Few-shots}} \\Read the following passage and then write Python code to answer the question:\\
        \textbf{\#\#\#Passage:} \clb{\texttt{text + table}}\\
        \textbf{\#\#\#Question:} \clb{ask question?}\\
        \textbf{\#\#\#Python}\\
};
        \addHeader{p}{ph}{magenta}{red}{Few-shot prompt: FinQA and TATQA}
\node (q) [miniprompt=lightred,right=of m, yshift=-1.4em, xshift=1em] {\\\phantom{hello} \\\clm{\textbf{Few-shots}} \\Read the following text and table, and then answer the last question by writing a Python code:\\
        \textbf{\#\#\#Passage:} \clb{\texttt{text + table}}\\
        \textbf{\#\#\#Questions:} \clb{A series of questions of a conversation?}\\
        \textbf{\#\#\#Last Question:} \clb{Last question of the conversation?}\\
        \textbf{\#\#\#Python}\\
};
        \addHeader{q}{qh}{darkgreen}{green}{Few-shot prompt: ConvFinQA}
\end{tikzpicture}
\end{center}
\caption{Few-shot prompts. \label{fig:few_shot_prompts}}
\end{figure*}

\begin{figure*}
\begin{center}
\begin{tikzpicture}[node distance=5mm]
\node (p) [miniprompt=cyan, yshift=-1em] {\\\phantom{hello} \\Read the following passage and then write Python code to answer the question:\\
        \textbf{\#\#\#Passage:} \clb{\texttt{text + table}}\\
        \textbf{\#\#\#Question:} \clb{ask question?}\\
	\textbf{\#\#\#Instructions:} The final answer must be stored in the Python variable "ans" and comments must begin with character "\#".\\
        \textbf{\#\#\#Python}\\
	\par\noindent\rule{\textwidth}{1pt}\\
	\clb{\texttt{\gpt code}} \\
	\textbf{\#\#\#EndPython}
};
        \addHeader{p}{ph}{magenta}{red}{Fine-tuning prompt-completion pairs: FinQA and TATQA}
\node (q) [miniprompt=lightred,right=of m, yshift=-1.4em, xshift=1em] {\\\phantom{hello} \\Read the following text and table, and then answer the last question by writing a Python code:\\
        \textbf{\#\#\#Passage:} \clb{\texttt{text + table}}\\
        \textbf{\#\#\#Questions:} \clb{A series of questions of a conversation?}\\
        \textbf{\#\#\#Last Question:} \clb{Last question of the conversation?}\\
	\textbf{\#\#\#Instructions:} The final answer must be stored in the Python variable "ans" and comments must begin with character "\#".\\
        \textbf{\#\#\#Python}\\
	\par\noindent\rule{\textwidth}{1pt}\\
	\clb{\texttt{\gpt code}} \\
	\textbf{\#\#\#EndPython}
};
        \addHeader{q}{qh}{darkgreen}{green}{Fine-tuning prompt-completion pairs: ConvFinQA}
\end{tikzpicture}
\end{center}
\caption{Fine-tuning prompt-completion pairs. \label{fig:fine_tuning_prompts}}
\end{figure*}
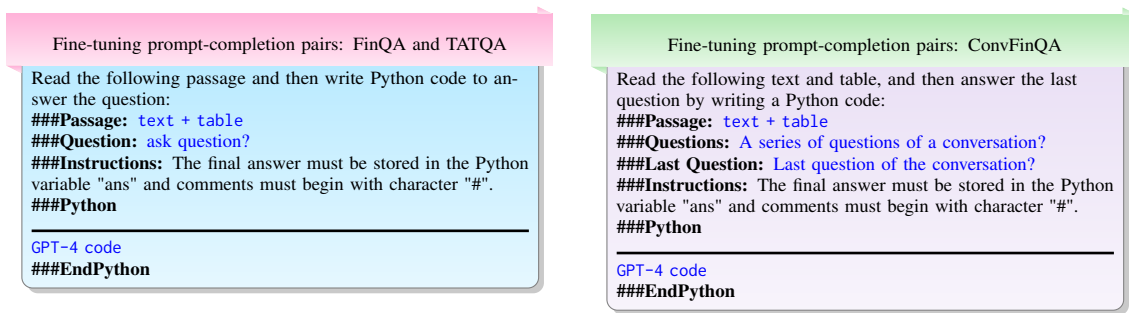

The zero-shot, few-shot and fine-tuning prompts for all datasets are given in Figure \ref{fig:zero_shot_prompts}, Figure \ref{fig:few_shot_prompts} and Figure \ref{fig:fine_tuning_prompts} respectively. We use the chat template that comes with the model's tokenizer for formatting prompt and completion into model specific formats. The formatted prompts and completions are then used during supervised fine-tuning of models and their evaluations.       

\section{\gpt for Student Model Assessment}\label{app:gpt_judge}
We use \gpt to quantify the LLMs conceptual understanding and entity extraction capabilities. In performing this analysis, \gpt is prompted to identify if the student code's concept or extracted entities are correct or not. The zero-shot prompt for evaluating these capabilities is given in Figure \ref{fig:gpt_judge_concept} and Figure \ref{fig:gpt_judge_entity}.    

\begin{figure}
\begin{center}
\begin{tikzpicture}[node distance=12mm]
\node (pp) [miniprompt=cyan, yshift=-1em] {\\\phantom{hello} \\You are an AI assistant. You will be given the definition of an evaluation metric for assessing the financial concept understanding demonstrated by the provided student code. The assessment is based on the provided question and the gold code that provides the true concept. Please note that the student code can be in a very different format and the format difference should be ignored. Please ignore the values of the required entities in your assessment. Entity extraction is a different skill that is not  relevant to assess concept understanding. Your job is to compute an accurate evaluation score using the provided evaluation metric. Make sure to explain your answer.\\
Concept understanding measures how well the student model code demonstrates an understanding of the financial concept illustrated by the gold code. Consider whether the student code is trying to compute the required entity and is it talking about entities relevant to the required computation. Given the student code, the gold code and the question, score the concept understanding demonstrated by the student code between one to five stars using the following rating scale:\\
\textbf{\texttt{One star:}} the student code demonstrates no understanding of the concept to be calculated\\
\textbf{\texttt{Two stars:}} the student code demonstrates limited understanding of the required concept \\
\textbf{\texttt{Three stars:}} the student code demonstrates partial understanding of the required concept\\
\textbf{\texttt{Four stars:}} the student code mostly demonstrates the understanding of the concept illustrated by the gold code but there are minor issues\\
\textbf{\texttt{Five stars:}} the student code demonstrates perfect understanding of the concept illustrated by the gold code.\\
This rating value should always be an integer between 1 and 5. So the rating produced should be 1 or 2 or 3 or 4 or 5. The result should strictly be written in the following format: \{'Explanation': [Think step by step and explain the reason in detail for the rating. Step 1: Analyse the gold code, Step 2: Analyse the student code, Step 3: Evaluate the student code by comparing with the gold code. Step 4: Provide the final rating with a detailed justification.], 'Star rating': [int]\}"
\\
	\textbf{Question:} \clb{\texttt{question?}}\\
	\textbf{Gold code:} \clb{\texttt{Teacher's generated code}}\\
	\textbf{Student generated code:} \clb{Student's generated code}
};
        \addHeader{pp}{pph}{magenta}{red}{Concept evaluation prompt}
\end{tikzpicture}
\end{center}
\caption{Zero-shot prompt for concept assessment using \gpt. \label{fig:gpt_judge_concept}}
\end{figure}

\begin{figure}
\begin{center}
\begin{tikzpicture}[node distance=12mm]
\node (p) [miniprompt=cyan, yshift=-1em] {\\\phantom{hello} \\Based on the question and the gold answer, determine if the student has correctly extracted all the relevant entities. Strictly ensure that the entity values are exactly matching.\\
	\textbf{Question:} \clb{\texttt{question?}}\\
	\textbf{Gold code:} \clb{\texttt{Teacher's generated code}}\\
	\textbf{Student generated code:} \clb{Student's generated code}
};
        \addHeader{p}{ph}{magenta}{red}{Entity extraction evaluation prompt}
\end{tikzpicture}
\end{center}
\caption{Zero-shot prompt for entity extraction assessment using \gpt. \label{fig:gpt_judge_entity}}
\end{figure}

\newcommand{\distTikz}[1]{
\begin{figure}
\centering
\begin{tikzpicture}[scale=0.7]
  \begin{axis}[
    ybar,
    bar width=0.75cm,
    width=10cm,
    height=5cm,
    xticklabels from table={\datatable}{indexRating},
    xticklabel style={font=\small, align=center, text width=4.5cm, inner sep=0pt},
    xtick=data,
    enlarge x limits=0.5,
    ymin=0,
    ymax=100, 
    ylabel={\footnotesize Percentage},
    legend style={at={(0.475, 1.25)},
    anchor=north,legend columns=-1},
    ymajorgrids=true,
    grid style={dashed,gray!60},
    nodes near coords,
    every node near coord/.append style={font=\small},
    nodes near coords align={vertical},
  ]

  \addplot[fill=red!50!magenta!50, bar shift=-1.5cm] table[x expr=\coordindex, y expr=\thisrow{r1}] {\datatable};
  \addplot[fill=red!50!orange!50, bar shift=-0.75cm] table[x expr=\coordindex, y expr=\thisrow{r2}] {\datatable};
  \addplot[fill=darkgreen!50!yellow!50, bar shift=0cm] table[x expr=\coordindex, y expr=\thisrow{r3}] {\datatable};
  \addplot[fill=violet!50!blue!50, bar shift=0.75cm] table[x expr=\coordindex, y expr=\thisrow{r4}] {\datatable};
  \addplot[fill=blue!50!cyan!50, bar shift=1.5cm] table[x expr=\coordindex, y expr=\thisrow{r5}] {\datatable};

  \legend{\footnotesize rating-1, \footnotesize rating-2, \footnotesize rating-3, \footnotesize rating-4, \footnotesize rating-5}
  \end{axis}
\end{tikzpicture}
\caption{#1}
\end{figure}
}

\pgfplotstableread[col sep=space]{
indexRating	r1	r2	r3 	r4 	r5	
{Base model}	28.6	15.0	16.2	12.3	28.0	
{Fine-tuned model (epoch-1)}	6.6	6.1	6.3	4.1	77.0	
}\datatable
\distTikz{\label{fig:gpt_rating_orca7b}Concept rating for \Orca-7B by \gpt.}

\pgfplotstableread[col sep=space]{
indexRating	r1	r2	r3 	r4 	r5	
{Base model}	7.7	 14.9	 15.0	 16.0	 46.5	
{Fine-tuned model (epoch-1)}	4.2	 4.6	 5.4	 6.1	 79.7	
}\datatable
\distTikz{\label{fig:gpt_rating_orca13b}Concept rating for \Orca-13B by \gpt.}

\pgfplotstableread[col sep=space]{
indexRating	r1	r2	r3 	r4 	r5	
{Base model}	6.4	 12.1	 13.7	 14.7	 53.0	
{Fine-tuned model (epoch-1)}	2.8	 4.0	 5.4	 3.6	 84.2	
}\datatable
\distTikz{\label{fig:gpt_rating_mistral}Concept rating for \Mistral 7B by \gpt.}

\pgfplotstableread[col sep=space]{
indexRating	r1	r2	r3 	r4 	r5	
{Base model}	7.2	 9.1	 6.8	 7.7	 69.2	
{Fine-tuned model (epoch-1)}	3.7	 5.2	 5.3	 2.8	 82.9	
}\datatable
\distTikz{\label{fig:gpt_rating_phimn}Concept rating for \Phimn by \gpt.}

\pgfplotstableread[col sep=space]{
indexRating	r1	r2	r3 	r4 	r5	
{Base model}	4.7	 6.6	 6.2	 3.3	 79.2	
{Fine-tuned model (epoch-1)} 	2.0	 3.6	 4.8	 5.1	 84.5	
}\datatable
\distTikz{\label{fig:gpt_rating_phimd}Concept rating for \Phimd by \gpt.}

\section{Examples}\label{app:examples}
We illustrate with examples how the student models \Phimn and \Orca-7B improve in various areas through training. These examples display the student models' responses at the base checkpoint, i.e., checkpoint-0, and after training for epoch 1, i.e., checkpoint-6000. The example shown in Figure \ref{fig:orca_code_gen} demonstrates that the code generation capabilities of the \Orca-7B model improve significantly after training for epoch 1. The example provided in Figure \ref{fig:orca_entity_extraction} illustrates that the \Orca-7B model's entity extraction capabilities improve with training.

In some instances, the base \Phimn model fails to generate a structured response and misses concept generation, as shown in Figure \ref{fig:phi_concept_not}. However, this is corrected by training the model for an epoch. The example in Figure \ref{fig:phi_des_var} demonstrates a case where the entity names are not descriptive at the base checkpoint, but this improves with model training. In general, the improvement in complex concepts does not occur with training the model for an epoch, and this observation is true for the \Phimn model as well, as seen in Figure \ref{fig:phi_complex_concept}.

\begin{figure*}[htb]
\begin{center}
\begin{tikzpicture}[node distance=12mm]
        \node (n1) [example_prompt] {
\qaid{IP/2005/page\_19.pdf-1}

\qabox{gray}{
\textbf{Passage:} item 7 . management 2019s discussion and analysis of financial condition and results of operations executive summary international paper 2019s operating results in 2005 were strongly impacted by significantly higher costs for energy $\ldots$\\
in millions | 2005 | 2004 | 2003\\
industry segment operating profits | \$ 1923 | \$ 2040 | \$ 1734\\
corporate items | -597 ( 597 ) | -469 ( 469 ) | -466 ( 466 )\\
corporate special items* | -147 ( 147 ) | -142 ( 142 ) | -281 ( 281 )\\
interest expense net | -593 ( 593 ) | -710 ( 710 ) | -705 ( 705 )\\
accounting changes | 2013 | 2013 | -13 ( 13 )\\
net earnings ( loss ) | \$ 1100 | \$ -35 ( 35 ) | \$ 302\\ 
\textbf{Question:} what was the change in industry segment operating profits between 2004 and 2005? 
}

\gptbox{blue}{
\textbf{Gold Program:} subtract(1923, 2040)\\
\textbf{Gold Answer:} -117.0 
}

\gptbox{darkgreen}{
\textbf{\Orca-7B Checkpoint-0 Code:}\\
To answer the question, we need to find the industry segment operating profits for both 2004 and 2005 and then calculate the change between the two years.\\
First, let's find the industry segment operating profits for 2004 and 2005:\\
2004:\\
Industry segment operating profits = \$2040 million\\
2005:\\
Industry segment operating profits = \$1923 million\\
Now, let's calculate the change in industry segment operating profits between 2004 and 2005:\\
Change = Industry segment operating profits in 2005 - Industry segment operating profits in 2004\\
Change = \$1923 million - \$2040 million\\
Change = -\$117 million\\
So, the change in industry segment operating profits between 2004 and 2005 was -\$117 million.\\
Final answer: -\$117 million\\
\textbf{\Orca-7B Checkpoint-0 Executed Answer:} \texttt{Null}
}

\gptbox{violet}{
\textbf{\Orca-7B Checkpoint-6000 Code:}\\
\# Calculate: change\_in\_industry\_segment\_operating\_profits = industry\_segment\_operating\_profits\_2005 - industry\_segment\_operating\_profits\_2004\\
industry\_segment\_operating\_profits\_2004 = 2040\\
industry\_segment\_operating\_profits\_2005 = 1923\\
change\_in\_industry\_segment\_operating\_profits = industry\_segment\_operating\_profits\_2005 - industry\_segment\_operating\_profits\_2004\\
ans = change\_in\_industry\_segment\_operating\_profits\\
\#\#\#Final Answer: change\_in\_industry\_segment\_operating\_profits = 117\\
\textbf{\Orca-7B Checkpoint-6000 Executed Answer:} -117.0
}
};
\end{tikzpicture}
\end{center}
\caption{An example illustrating the code generation capabilities of the fine-tuned \Orca-7B model. \label{fig:orca_code_gen}}
\end{figure*}

\begin{figure*}[htb]
\begin{center}
\begin{tikzpicture}[node distance=12mm]
        \node (n1) [example_prompt] {
\qaid{IP/2005/page\_35.pdf-4}

\qabox{gray}{
\textbf{Passage:} contractual obligations for future payments under existing debt and lease commitments and purchase obli- gations at december 31 , 2005 , were as follows : in millions 2006 2007 2008 2009 2010 thereafter . ( a ) the 2006 amount includes \$ 2.4 billion for contracts made in the ordinary course of business to purchase pulpwood , logs and wood chips . the majority of our other purchase \\
in millions | 2006 | 2007 | 2008 | 2009 | 2010 | thereafter\\
total debt | \$ 1181 | \$ 570 | \$ 308 | \$ 2330 | \$ 1534 | \$ 6281\\
lease obligations | 172 | 144 | 119 | 76 | 63 | 138\\
purchase obligations ( a ) | 3264 | 393 | 280 | 240 | 204 | 1238\\
total | \$ 4617 | \$ 1107 | \$ 707 | \$ 2646 | \$ 1801 | \$ 7657\\
\textbf{Question:} what was the percent of the total contractual obligations for future payments for total debt in 2005? 
}

\gptbox{blue}{
\textbf{Gold Program:} divide(1181, 4617) \\
\textbf{Gold Answer:}0.25579 
}

\gptbox{darkgreen}{
\textbf{\Orca-7B Checkpoint-0 Code:}\\
To answer the question, we need to find the total contractual obligations for future payments for total debt in 2005 and then calculate the percentage of that amount.\\
From the passage, we can see that the total debt in 2005 was \$1181 million.\\
Now, let's calculate the percentage:\\
Percentage = (Total debt in 2005 / Total contractual obligations for future payments for total debt in 2005) * 100\\
Percentage = (\$1181 million / Total contractual obligations for future payments for total debt in 2005) * 100\\
Percentage = (\$1181 / ?) * 100\\
We don't have the total contractual obligations for future payments for total debt in 2005, so we cannot calculate the percentage.\\
\textbf{\Orca-7B Checkpoint-0 Executed Answer:} \texttt{Null}
}
\gptbox{violet}{
\textbf{\Orca-7B Checkpoint-6000 Code:}\\
\# Calculate: percent\_of\_total\_contractual\_obligations\_for\_future\_payments\_for\_total\_debt\_2005 = ( total\_debt\_2005 / total\_contractual\_obligations\_2005 ) * 100\\
total\_debt\_2005 = 1181\\
total\_contractual\_obligations\_2005 = 4617\\
percent\_of\_total\_contractual\_obligations\_for\_future\_payments\_for\_total\_debt\_2005 = ( total\_debt\_2005 / total\_contractual\_obligations\_2005 ) * 100\\
ans = percent\_of\_total\_contractual\_obligations\_for\_future\_payments\_for\_total\_debt\_2005 \\
\#\#\#Final Answer: 23.6\%\\
\textbf{\Orca-7B Checkpoint-6000 Executed Answer:} 25.57938 
}
};
\end{tikzpicture}
\end{center}
\caption{An example illustrating enhanced entity extraction capability with training \Orca-7B model. \label{fig:orca_entity_extraction}}
\end{figure*}

\begin{figure*}[htb]
\begin{center}
\begin{tikzpicture}[node distance=12mm]
        \node (n1) [example_prompt] {
\qaid{IP/2007/page\_31.pdf-4}
\qabox{gray}{
\textbf{Passage:} tissue pulp due to strong market demand, particularly from asia. average sales price realizations improved significantly in 2007 $\ldots$ \\
in millions | 2007 | 2006 | 2005\\
sales | \$ 5245 | \$ 4925 | \$ 4625\\
operating profit | \$ 501 | \$ 399 | \$ 219 \\
\textbf{Question:} in 2005 what was industrial packaging the profit margin? 
}

\gptbox{blue}{
\textbf{Gold Program:} divide(219, 4625)\\
\textbf{Gold Answer:} 0.04735 
}

\gptbox{darkgreen}{
\textbf{\Phimn Checkpoint-0 Code:}\\
\# Read the relevant data from the passage\\
profit\_margin\_2005 = 219  \# The profit margin for industrial packaging in 2005 is given as \$219 million\\

\# Store the answer in the variable "ans"\\
ans = profit\_margin\_2005\\ 
\textbf{\Phimn Checkpoint-0 Executed Answer:} 219.0 
}

\gptbox{violet}{
\textbf{\Phimn Checkpoint-6000 Code:}\\
\# Calculate: profit\_margin\_2005 = operating\_profit\_2005 / sales\_2005 \\
operating\_profit\_2005 = 219 \\
sales\_2005 = 4625 \\
profit\_margin\_2005 = operating\_profit\_2005 / sales\_2005 \\
ans = profit\_margin\_2005 \\
\#\#\#End Python\\
\textbf{\Phimn Checkpoint-6000 Executed Answer:} 0.04735
}
};
\end{tikzpicture}
\end{center}
\caption{An example where the concept is not clearly stated (\gpt concept rating 1) at checkpoint-0 that improves to \gpt concept rating 5 after training \Phimn for 1 epoch. \label{fig:phi_concept_not}}
\end{figure*}

\begin{figure*}[htb]
\begin{center}
\begin{tikzpicture}[node distance=12mm]
        \node (n1) [example_prompt] {
\qaid{SPGI/2018/page\_74.pdf-1}
\qabox{gray}{
\textbf{Passage:} 72 s\&p global 2018 annual report. for securities that are quoted in active markets, the trustee/ custodian determines fair value by applying $\ldots$ \\
( in millions ) | level 3\\
balance as of december 31 2017 | \$ 39\\
purchases | 2014\\
distributions | -2 ( 2 )\\
gain ( loss ) | 2\\
balance as of december 31 2018 | \$ 39\\
\textbf{Question:} what was the ratio of the pension trust assets for 2017 to 2018? 
}

\gptbox{blue}{
\textbf{Gold Program:} divide(480, 415)\\
\textbf{Gold Answer:} 1.15663 
}

\gptbox{darkgreen}{
\textbf{\Phimn Checkpoint-0 Code:}\\
\# First, we extract the pension trust assets for 2017 and 2018 from the passage.\\
\\
\# Pension trust assets for 2017\\
assets\_2017 = 480  \# in millions\\
\\
\# Pension trust assets for 2018\\
assets\_2018 = 1739  \# in millions\\
\\
\# Now, we calculate the ratio of the pension trust assets for 2017 to 2018.\\
\\
\# Calculate the ratio\\
ratio = assets\_2017 / assets\_2018\\
\\
\# Store the result in the variable "ans"\\
ans = ratio \\ 
\textbf{\Phimn Checkpoint-0 Executed Answer:} 0.27602 
}

\gptbox{violet}{
\textbf{\Phimn Checkpoint-6000 Code:}\\
\# Calculate: ratio\_of\_pension\_trust\_assets = pension\_trust\_assets\_2017 / pension\_trust\_assets\_2018\\
pension\_trust\_assets\_2017 = 1739\\
pension\_trust\_assets\_2018 = 1572\\
ratio\_of\_pension\_trust\_assets = pension\_trust\_assets\_2017 / pension\_trust\_assets\_2018\\
ans = ratio\_of\_pension\_trust\_assets \\
\#\#\#End Python\\
\textbf{\Phimn Checkpoint-6000 Executed Answer:} 1.10623 
}
};
\end{tikzpicture}
\end{center}
\caption{An example where the descriptive response (\gpt concept rating 5) at checkpoint-0 follows the structured format after training \Phimn for 1 epoch. \label{fig:phi_des_var}}
\end{figure*}

\begin{figure*}[htb]
\begin{center}
\begin{tikzpicture}[node distance=12mm]
        \node (n1) [example_prompt] {
\qaid{APD/2019/page\_31.pdf-2}
\qabox{gray}{
\textbf{Passage:} part ii item 5 . market for registrant 2019s common equity , related stockholder matters , and issuer purchases of equity securities our common stock $\ldots$ \\
-  | 2019 | 2018\\
first quarter | \$ 1.10 | \$ .95\\
second quarter | 1.16 | 1.10\\
third quarter | 1.16 | 1.10\\
fourth quarter | 1.16 | 1.10\\
total | \$ 4.58 | \$ 4.25\\
\textbf{Question:} what is the increase observed in the first quarter dividend percentage concerning the total value , during 2018 and 2019? 
}

\gptbox{blue}{
\textbf{Gold Program:} divide(1.10, 4.58), divide(.95, 4.25), subtract(\#0, \#1)\\
\textbf{Gold Answer:} 0.01665 
}

\gptbox{darkgreen}{
\textbf{\Phimn Checkpoint-0 Code:}\\
\# First, we need to extract the dividend percentages for the first quarter of both 2018 and 2019 from the given data.\\
\\
\# Extracting the first quarter dividend percentage for 2018\\
first\_quarter\_2018\_dividend\_percentage = 1.10  \# Given in the passage\\
\\
\# Extracting the first quarter dividend percentage for 2019\\
first\_quarter\_2019\_dividend\_percentage = 1.16  \# Given in the passage\\
\\
\# Calculating the increase in the first quarter dividend percentage from 2018 to 2019\\
increase\_in\_dividend\_percentage = first\_quarter\_2019\_dividend\_percentage - first\_quarter\_2018\_dividend\_percentage\\
\\
\# Storing the final answer in the variable "ans"\\
ans = increase\_in\_dividend\_percentage \\
\textbf{\Phimn Checkpoint-0 Executed Answer:} 0.05999
}

\gptbox{violet}{
\textbf{\Phimn Checkpoint-6000 Code:}\\
\# Calculate: increase\_in\_first\_quarter\_dividend\_percentage = ( first\_quarter\_dividend\_2019 - first\_quarter\_dividend\_2018 ) / first\_quarter\_dividend\_2018\\
first\_quarter\_dividend\_2018 = 0.95\\
first\_quarter\_dividend\_2019 = 1.10\\
increase\_in\_first\_quarter\_dividend\_percentage = ( first\_quarter\_dividend\_2019 - first\_quarter\_dividend\_2018 ) / first\_quarter\_dividend\_2018\\
ans = increase\_in\_first\_quarter\_dividend\_percentage \\
\#\#\#End Python\\
\textbf{\Phimn Checkpoint-6000 Executed Answer:} 0.15789 
}
};
\end{tikzpicture}
\end{center}
\caption{An example where the complex concept with \gpt concept rating 2 remains unchanged after training \Phimn for 1 epoch. \label{fig:phi_complex_concept}}
\end{figure*}

\section{Few-shots for datasets: FinQA, ConvFinQA and TATQA}\label{app:few_shots}
We use few-shot prompting for code generation by the teacher model (\gpt). For all datasets, we use 4-shots in our experiments and these 4-shots for FinQA, ConvFinQA and TATQA datasets are listed in Figure \ref{fig:few_shot_finqa}, Figure \ref{fig:few_shot_convfinqa} and Figure \ref{fig:few_shot_tatqa} respectively. 
\begin{figure*}[htb]
\begin{center}
\begin{tikzpicture}[node distance=12mm]
        \node (n1) [example_prompt] {
\qabox{gray}{
Read the following passage and then write python code to answer the question\\
\textbf{\#\#\#Passage:}  
( in millions ) | dec 282013 | dec 292012\\
available-for-sale investments | \$ 18086 | \$ 14001\\
cash | 854 | 593\\
equity method investments | 1038 | 992\\
loans receivable | 1072 | 979\\
non-marketable cost method investments | 1270 | 1202\\ 
reverse repurchase agreements | 800 | 2850\\
trading assets | 8441 | 5685\\
total cash and investments | \$ 31561 | \$ 26302\\
\textbf{\#\#\#Question:} what percentage of total cash and investments as of dec . 29 2012 was comprised of available-for-sale investments?\\
\textbf{\#\#\#Python}\\  
\#Calculate: percent\_available\_for\_sale\_investments\_dec\_29\_2012 = available\_for\_sale\_investments\_dec\_29\_2012 / total\_cash\_and\_investments\_dec\_29\_2012\\
available\_for\_sale\_investments\_dec\_29\_2012 = 14001\\
total\_cash\_and\_investments\_dec\_29\_2012 = 26302\\
percent\_available\_for\_sale\_investments\_dec\_29\_2012 = available\_for\_sale\_investments\_dec\_29\_2012 / total\_cash\_and\_investments\_dec\_29\_2012\\
ans = percent\_available\_for\_sale\_investments\_dec\_29\_2012\\
\textbf{\#\#\#End Python</s>}
}
\qabox{gray}{
Read the following passage and then write python code to answer the question\\
\textbf{\#\#\#Passage:}
the chart shows that the firm posted market risk 2013 related gains on 248 out of 261 days in this period , with 12 days exceeding \$ 210 million .
december 31 ( in millions ) | 1 basis point increase in jpmorgan chase 2019s credit spread\\
2010 | \$ 35\\
2009 | \$ 39\\
\textbf{\#\#\#Question:} on what percent of trading days were there market gains above \$ 210 million?\\
\textbf{\#\#\#Python}\\
\#Calculate: percent\_days\_with\_market\_gains\_above\_210\_million = days\_with\_market\_gains\_above\_210\_million / total\_trading\_days\\
days\_with\_market\_gains\_above\_210\_million = 12\\
total\_trading\_days = 261\\
percent\_days\_with\_market\_gains\_above\_210\_million = days\_with\_market\_gains\_above\_210\_million / total\_trading\_days\\
ans = percent\_days\_with\_market\_gains\_above\_210\_million\\
\textbf{\#\#\#End Python</s>}
}
\qabox{gray}{
Read the following passage and then write python code to answer the question\\
\textbf{\#\#\#Passage:}
american tower corporation and subsidiaries notes to consolidated financial statements ( 3 ) consists of customer-related intangibles of approximately \$ 75.0 million and network location intangibles of approximately \$ 72.7 million . the customer-related intangibles and network location intangibles are being amortized on a straight-line basis over periods of up to 20 years.\\
- | preliminary purchase price allocation\\
current assets | \$ 8763\\
$\ldots$ \\
fair value of net assets acquired | \$ 57536\\
goodwill ( 2 ) | 5998\\
\textbf{\#\#\#Question:} for acquired customer-related and network location intangibles , what is the expected annual amortization expenses , in millions?\\
\textbf{\#\#\#Python}\\
\#Calculate: amortization\_expenses = ( customer\_related\_intangibles + network\_location\_intangibles ) / straight\_line\_basis\\
customer\_related\_intangibles = 75\\
network\_location\_intangibles = 72.7\\
straight\_line\_basis = 20\\
amortization\_expenses = ( customer\_related\_intangibles + network\_location\_intangibles ) / straight\_line\_basis\\
ans = amortization\_expenses\\
\textbf{\#\#\#End Python</s>}
}
\qabox{gray}{
Read the following passage and then write python code to answer the question\\
\textbf{\#\#\#Passage:}
the aggregate commitment under the liquidity asset purchase agreements was approximately \$ 23.59 billion and \$ 28.37 billion at december 31 , 2008 and 2007 , respectively .\\
( dollars in billions ) | 2008 amount | 2008 percent of total conduit assets | 2008 amount | percent of total conduit assets\\
united states | \$ 11.09 | 46\% ( 46 \% ) | \$ 12.14 | 42\% ( 42 \% )\\
australia | 4.30 | 17 | 6.10 | 21\\
$\ldots$ \\
greece | 0.27 | 1 | 0.31 | 1\\
other | 1.01 | 5 | 1.26 | 5\\
total conduit assets | \$ 23.89 | 100\% ( 100\% ) | \$ 28.76 | 100\% ( 100\% )\\
\textbf{\#\#\#Question:} what is percentage change in total conduit asset from 2007 to 2008?\\
\textbf{\#\#\#Python}\\
\#Calculate: percent\_change\_in\_total\_conduit\_assets = ( total\_conduit\_assets\_2008 - total\_conduit\_assets\_2007 ) / total\_conduit\_assets\_2007\\
total\_conduit\_assets\_2007 = 28.76\\
total\_conduit\_assets\_2008 = 23.89\\
net\_change\_in\_total\_conduit\_assets = total\_conduit\_assets\_2008 - total\_conduit\_assets\_2007\\
percent\_change\_in\_total\_conduit\_assets = net\_change\_in\_total\_conduit\_assets / total\_conduit\_assets\_2007\\
ans = percent\_change\_in\_total\_conduit\_assets\\
\textbf{\#\#\#End Python</s>}
}
};
\end{tikzpicture}
\end{center}
\caption{Few shots for FinQA.}
\label{fig:few_shot_finqa}
\end{figure*}

\begin{figure*}[htb]
\begin{center}
\begin{tikzpicture}[node distance=12mm]
        \node (n1) [example_prompt] {
\qabox{gray}{
Read the following text and table, and then answer the last question in a series of questions:\\
\textbf{\#\#\#Passage:}\\
- | shares available for awards | shares subject to outstanding awards\\
2009 global incentive plan | 2322450 | 2530454\\
2004 stock incentive plan | - | 5923147\\
\textbf{\#\#\#Questions:} how many shares are subject to outstanding awards is under the 2009 global incentive plan? what about under the 2004 stock incentive plan? how many total shares are subject to outstanding awards? what about under the 2004 stock incentive plan?\\
\textbf{\#\#\#Last Question:} what proportion does this represent?\\
\textbf{\#\#\#Python}\\  
\# Calculate: shares\_outstanding\_awards\_2009\_global\_incentive\_plan / ( shares\_outstanding\_awards\_2009\_global\_incentive\_plan + shares\_outstanding\_awards\_2004\_stock\_incentive\_plan ) \\
shares\_outstanding\_awards\_2009\_global\_incentive\_plan = 2530454\\
shares\_outstanding\_awards\_2004\_stock\_incentive\_plan = 5923147\\
total\_shares\_outstanding\_awards = shares\_outstanding\_awards\_2009\_global\_incentive\_plan + \\ shares\_outstanding\_awards\_2004\_stock\_incentive\_plan\\
proportion = shares\_outstanding\_awards\_2009\_global\_incentive\_plan / total\_shares\_outstanding\_awards\\
ans = proportion\\
\textbf{\#\#\#End Python</s>}
}
\qabox{gray}{
Read the following text and table, and then answer the last question in a series of questions:\\
\textbf{\#\#\#Passage:}
compensation expense the company recorded \$ 43 million , \$ 34 million , and \$ 44 million of expense related to stock awards for the years ended december 31 , 2015 , 2014 , and 2013 , respectively . \\
\textbf{\#\#\#Questions:} what is the compensation expense the company recorded in 2015? what about in 2014? what is the total compensation expense the company recorded in 2015 and 2014? what is the total expenses including 2013?\\
\textbf{\#\#\#Last Question:} what is the average for three years?\\
\textbf{\#\#\#Python}\\
\# Calculate: average\_for\_three\_years = ( compensation\_expense\_2015 + compensation\_expense\_2014 + compensation\_expense\_2013 ) /  3\\
compensation\_expense\_2015 = 43\\
compensation\_expense\_2014 = 34\\
compensation\_expense\_2013 = 44\\
total\_compensation\_expense = compensation\_expense\_2015 + compensation\_expense\_2014 + compensation\_expense\_2013\\
average\_for\_three\_years = total\_compensation\_expense / 3\\
ans = average\_for\_three\_years\\
\textbf{\#\#\#End Python</s>}
}
\qabox{gray}{
Read the following text and table, and then answer the last question in a series of questions:\\
\textbf{\#\#\#Passage:}
the net loss on disposal of those assets was \$ 344000 for 2005 and \$ 43000 for 2004 . \\
\textbf{\#\#\#Questions:} what was the net loss on disposal of assets in 2005? what was the value in 2004? what was the change in value?\\
\textbf{\#\#\#Last Question:} what was the percent change?\\
\textbf{\#\#\#Python}\\
\# Calculate: percent\_change = ( net\_loss\_on\_disposal\_of\_assets\_2005 - net\_loss\_on\_disposal\_of\_assets\_2004 ) / net\_loss\_on\_disposal\_of\_assets\_2004\\
net\_loss\_on\_disposal\_of\_assets\_2005 = 344000\\
net\_loss\_on\_disposal\_of\_assets\_2004 = 43000\\
net\_change\_in\_value = net\_loss\_on\_disposal\_of\_assets\_2005 - net\_loss\_on\_disposal\_of\_assets\_2004\\
percent\_change = net\_change\_in\_value / net\_loss\_on\_disposal\_of\_assets\_2004\\
ans = percent\_change\\
\textbf{\#\#\#End Python</s>}
}
\qabox{gray}{
Read the following text and table, and then answer the last question in a series of questions:\\
\textbf{\#\#\#Passage:}
location | operations conducted | approximatesquare feet | leaseexpirationdates\\
dublin ireland | global supply chain distribution and administration offices | 160000 | owned\\
athlone ireland | commercial research and development manufacturing | 80000 | owned\\
bogart georgia | commercial research and development manufacturing | 70000 | owned\\
smithfield rhode island | commercial research and development manufacturing | 67000 | owned\\
\textbf{\#\#\#Questions:} what is the square feet of the owned global supply chain distribution and administration offices? what is the square feet of the owned commercial research and development manufacturing? what is the sum of those values? what is the total sum including square feet of commercial research and development manufacturing in bogart, georgia? what is the total sum including square feet of commercial research and development manufacturing in smithfield, rhode island?\\
\textbf{\#\#\#Last Question:} what is the total sum of square feet owned?\\
\textbf{\#\#\#Python}\\
\# Calculate: owned\_global\_supply\_chain\_distribution\_dublin + commercial\_research\_and\_development\_manufacturing\_athlone + commercial\_research\_and\_development\_manufacturing\_bogart + commercial\_research\_and\_development\_manufacturing\_smithfield \\
owned\_global\_supply\_chain\_distribution\_dublin = 160000\\
commercial\_research\_and\_development\_manufacturing\_athlone = 80000\\
commercial\_research\_and\_development\_manufacturing\_bogart = 70000\\
commercial\_research\_and\_development\_manufacturing\_smithfield = 67000\\
total\_square\_feet\_owned = owned\_global\_supply\_chain\_distribution\_dublin + commercial\_research\_and\_development\_manufacturing\_athlone + commercial\_research\_and\_development\_manufacturing\_bogart + commercial\_research\_and\_development\_manufacturing\_smithfield \\
ans = total\_square\_feet\_owned\\
\textbf{\#\#\#End Python</s>}
}
};
\end{tikzpicture}
\end{center}
\caption{Few shots for ConvFinQA.}
\label{fig:few_shot_convfinqa}
\end{figure*}

\begin{figure*}[htb]
\begin{center}
\begin{tikzpicture}[node distance=12mm]
        \node (n1) [example_prompt] {
\qabox{gray}{
Read the following passage and then write python code to answer the question\\
\textbf{\#\#\#Passage:}  
17. Income Taxes\\
Income before income taxes for the Company’s domestic and foreign operations was as follows:\\
— | — | Years Ended June 30, | —\\
(\$ in millions) | 2019 | 2018 | 2017\\
Domestic | \$204.2 | \$140.3 | \$56.0\\
Foreign | 11.8 | 19.9 | 14.2\\
Income before income taxes | \$216.0 | \$160.2 | \$70.2\\
\textbf{\#\#\#Question:} What was the change in Foreign in 2019 from 2018?\\
\textbf{\#\#\#Python}\\  
\# Calculate: change\_in\_foreign = foreign\_in\_2019 - foreign\_in\_2018\\
foreign\_in\_2018 = 19.9\\
foreign\_in\_2019 = 11.8\\
ans = foreign\_in\_2019 - foreign\_in\_2018\\
\textbf{\#\#\#End Python</s>}
}
\qabox{gray}{
Read the following passage and then write python code to answer the question\\
\textbf{\#\#\#Passage:}
11 Intangible assets (continued)
(a) Intangible assets
RIGHTS AND LICENCES
Certain licences that NEXTDC possesses have an indefinite useful life and are carried at cost less impairment losses and are subject to impairment review at least annually and whenever there is an indication that it may be impaired.\\
$\ldots$ \\
— | Rights and licenses | Internally generated software | Software under development | Total\\
At 30 June 2019 | — | — | — | —\\
Cost | 13 | 12,961 | 16,284 | 29,259\\
Accumulated amortisation | - | -5,580 | - | -5,580\\
At 30 June 2018 | — | — | — | —\\
Cost | 104 | 9,555 | 6,509 | 16,168\\
Accumulated amortisation | -91 | -3,170 | - | -3,261\\
Net book amount | 13 | 6,385 | 6,509 | 12,907\\
\textbf{\#\#\#Question:} Which year have greater total accumulated amortisation?\\
\textbf{\#\#\#Python}\\
\# Calculate: find total accumulated amortization and choose year with maximum accumulated amortisation\\
total\_accumulated\_amortisation = \{'2019': 5580, '2018': 3261\}\\
ans = sorted(total\_accumulated\_amortisation.items(), key=lambda tup: tup[1], reverse=True)[0][0]\\
\textbf{\#\#\#End Python</s>}
}

\qabox{gray}{
Read the following passage and then write python code to answer the question\\
\textbf{\#\#\#Passage:}
The following table sets forth the breakdown of revenues by category and segment.\\
$\ldots$ \\ 
Year Ended December 31, | — | —\\
— | 2019 | 2018\\
Total Asia Pacific revenues | 6,490 | 7,859\\
Total Europe revenues | 36,898 | 36,149\\
Total North America revenues | 68,024 | 67,314\\
Total revenues | \$111,412 | 111,322\\
\textbf{\#\#\#Question:} In 2019, how many geographic regions have total revenues of more than \$20,000 thousand? \\
\textbf{\#\#\#Python}\\
\# Calculate: find locations with revenue more than  \$20,000 in a list and count\\
total\_revenues\_in\_all\_regions = \{'Asia Pacific': 6490, 'Europe': 36898, 'North America': 68024\}\\
regions\_have\_more\_than\_20000\_thousand\_total\_revenues = [k for k, v in total\_revenues\_in\_all\_regions.items() if v > 20000]\\
ans = len(regions\_have\_more\_than\_20000\_thousand\_total\_revenues)\\
\textbf{\#\#\#End Python</s>}
}

\qabox{gray}{
Read the following passage and then write python code to answer the question\\
\textbf{\#\#\#Passage:}
Effective Income Tax Rate
A reconciliation of the United States federal statutory income tax rate to our effective income tax rate is as follows:
In 2019 and 2018 we had pre-tax losses of \$19,573 and \$25,403, respectively, which are available for carry forward to offset future taxable income. \\
— | Year Ended | Year Ended\\
— | December 31, 2018 | December 31, 2019\\
United States federal statutory rate | 21.00\% | 21.00\%\\
State taxes, net of federal benefit | 1.99\% | -0.01\%\\
\textbf{\#\#\#Question:} What was the 2019 percentage change in pre-tax losses? \\
\textbf{\#\#\#Python}\\
\# Calculate: percentage\_change\_in\_pre\_tax\_losses = ( pre\_tax\_losses\_2019 - pre\_tax\_losses\_2018 ) / pre\_tax\_losses\_2018 * 100\\
pre\_tax\_losses\_2018 = 25403\\
pre\_tax\_losses\_2019 = 19573\\
net\_change = pre\_tax\_losses\_2019 - pre\_tax\_losses\_2018\\
ans = net\_change / pre\_tax\_losses\_2018 * 100\\
\textbf{\#\#\#End Python</s>}
}
};
\end{tikzpicture}
\end{center}
\caption{Few shots for TATQA.}
\label{fig:few_shot_tatqa}
\end{figure*}

\end{document}